\documentclass[11pt]{article}

\usepackage[final]{acl}

\usepackage{times}
\usepackage{latexsym}

\usepackage[T1]{fontenc}
\usepackage[utf8]{inputenc}

\usepackage{microtype}

\usepackage{inconsolata}

\usepackage{graphicx}

\title{%
    \textbf{Cross-Lingual Bias in Large Language Models:}\\[0.3em]
    \Large A Comparative Analysis of English and Swahili\\[1em]
}

\author{
Ruolei Zhang\thanks{Equal contribution.}
\quad
Teddy Njuguna\footnotemark[1]
\quad
Yue Feng \\
University of Birmingham \\
\texttt{rxz527@student.bham.ac.uk} \quad
\texttt{txn234@student.bham.ac.uk} \quad
\texttt{y.feng.6@bham.ac.uk}
}

\begin{document}
\maketitle
\begin{abstract}
Large language models are increasingly deployed in multilingual contexts, yet safety alignment and bias evaluation remain overwhelmingly English-centric. We investigate whether social biases generalise across languages by submitting 4,900 symmetric English--Swahili prompt pairs to GPT-5.2 and Gemini 2.5 Flash across nine demographic bias axes, yielding 19,600 completions evaluated for stereotype prevalence, sentiment, refusal behaviour, and cross-lingual semantic similarity. Our findings show that bias transforms rather than transfers: stereotype rates shifted by up to 12 percentage points on specific axes, Gemini's neutral-sentiment rate doubled in Swahili, and GPT-5.2 refused 169 prompts in English and zero in Swahili, consistent with refusal behaviour anchored to English-language surface forms at the behavioural level. Over 55\% of prompt pairs produced semantically dissimilar completions across both models. These reinforce the idea that English-only bias audits do not produce adequate coverage for multilingual deployment.
\end{abstract}

\section{Introduction}
 
Since the introduction of the Transformer architecture \cite{vaswani2017attention}, large language models such as GPT \citep{openai2024gpt} and Gemini \citep{team2023gemini} have become ubiquitous across diverse linguistic communities. Yet the safety infrastructure governing their behaviour remains overwhelmingly monolingual. Alignment procedures such as Reinforcement Learning from Human Feedback (RLHF) \citep{ouyang2022training} and red-teaming \citep{ganguli2022red} are largely developed and evaluated in English, creating a fundamental asymmetry: English-speaking users interact with extensively audited systems, while users in lower-resource languages encounter models whose safety behaviour is comparatively under-examined.
 
This asymmetry is rooted in training data. Common Crawl is approximately 42.6\% English and just 0.01\% Swahili \citep{commoncrawl2026stats}, a disparity of roughly 3,400-to-1. Swahili is spoken by over 100 million people and serves as a lingua franca across East Africa. If safety guardrails fail to generalise to Swahili, the situation is almost certainly worse for less represented languages.
 
One may assume that bias evaluated in English generalises to other languages. This rests on a premise: that multilingual models maintain language-invariant internal representations of social concepts. Prior work has shown that models learn structural abstractions that generalise across languages \citep{artetxe2020cross}, but whether this extends to social and cultural concepts, where training data distributions differ dramatically, remains an open question. This study provides a behavioural evaluation, comparing GPT-5.2 and Gemini 2.5 Flash across nine bias axes in English and Swahili using two complementary evaluation pipelines. We make three contributions:
 
\begin{enumerate}
    \item We provide the first systematic cross-lingual bias comparison between English and Swahili using open-ended generative prompts, enabling analysis of sentiment framing, hedging, and refusal behaviour.
    \item We provide behavioural evidence that GPT-5.2's refusal behaviour is anchored to English-language surface forms, with 169 refusals in English and zero in Swahili on identical prompt templates.
    \item We show that cross-lingual bias transforms rather than transfers: the same model produces different stereotype rates, sentiment profiles, and semantically dissimilar completions across languages.
\end{enumerate}

\subsection*{Objectives}
The concrete objectives of this project are:
\begin{enumerate}
    \item To construct a symmetric English--Swahili prompt dataset across nine bias axes using a descriptor-template methodology.
    \item To evaluate stereotype prevalence, sentiment framing, and refusal behaviour across GPT-5.2 and Gemini 2.5 Flash in both languages.
    \item To assess cross-lingual semantic consistency between English and Swahili completions using an automated judge pipeline.
    \item To investigate whether safety alignment mechanisms generalise across languages using refusal behaviour as a proxy.
    \item To provide mechanistic evidence for the source of cross-lingual divergence using the logit lens technique on an open-source model.
\end{enumerate}

\section{Literature Review}

\subsection{Bias Benchmarks}

\citet{parrish2022bbq} introduced BBQ, a dataset of over 58,000 examples testing whether models rely on stereotypes when context is ambiguous. BBQ is effective but limited to multiple-choice, which cannot capture sentiment framing, hedging, or refusal behaviour in open-ended generation. \citet{smith2022m} introduced HolisticBias, combining nearly 600 descriptor terms across 13 demographic axes with sentence templates to produce over 450,000 prompts, demonstrating that combinatorial descriptor-template approaches effectively uncover biases in generative models.

Beyond multiple-choice benchmarks, open-ended sentence completion has been used to directly evaluate harmful model behaviour. \citet{gehman-etal-2020-realtoxicityprompts} split 100,000 naturally occurring English sentences into prompt-continuation pairs and conditioned five pre-trained language models on the prompts, allowing each to generate up to 20 tokens. Even innocuous prompts elicited toxic completions at rates exceeding 48\%, and none of the detoxification methods they evaluated eliminated this behaviour entirely. However, their evaluation was conducted exclusively in English. This study adopts a comparable sentence-completion paradigm but introduces a cross-lingual dimension: each English prompt is paired with a Swahili equivalent sharing the same template and descriptor.

\subsection{Cross-Lingual Bias and Safety Alignment}

\citet{artetxe2020cross} demonstrated that monolingual transformer models learn internal abstractions that transfer across languages even without shared vocabularies, establishing that cross-lingual generalisation is an emergent property of deep language models. However, this transferability has been studied primarily in syntactic and semantic tasks; whether it extends to social and cultural dimensions remains unclear. \citet{xu2025survey} documents how unbalanced language proportions in multilingual training corpora produce misaligned internal representations, with different languages occupying differently aligned regions of a shared embedding space. This misalignment is most severe for low-resource languages, where limited training signal reduces the overlap between language-specific representations of the same concepts.

\citet{neplenbroek2024mbbq} extended BBQ to Dutch, Spanish, and Turkish (MBBQ), confirming greater bias in non-English languages even when controlling for cultural shifts. \citet{liang2025crosslang} tested explicit and implicit bias across five languages, finding substantial cross-lingual gaps and opposite patterns across bias types. \citet{cocu2025cross} showed that LLMs shift their framing of contested Romanian historical questions depending on prompt language, reproducing the dominant narratives in each language's training data. \citet{huijzer2025discrimination} found that bias in decision-making tasks transforms rather than simply transfers across languages. This extends to safety alignment: \citet{yong2023lowresource} demonstrated that translating harmful prompts into low-resource languages bypassed GPT-4's safety guardrails nearly 80\% of the time. Together, these factors exacerbate the risks for users interacting with LLMs primarily through low-resource languages, who face both weaker safety guardrails and more biased outputs.

\subsection{Research Gap}

Most cross-lingual bias studies focus on European or high-resource Asian languages, leaving African languages under-represented. Existing benchmarks are also frequently constrained to multiple-choice formats, limiting the analysis of generative behaviours such as sentiment framing, hedging, or refusal. \citet{gamboa2025social} reinforce the consensus that multilingual models exhibit social biases akin to monolingual counterparts, but raise critical open questions: How does this bias manifest? Do models exhibit symmetrical bias tendencies across linguistic contexts? This study addresses that gap by examining cross-lingual bias in English and Swahili across nine bias axes using open-ended generative prompts.
 
\section{Methodology}

\subsection{Experimental Design}
 
\begin{table}[ht]
\centering
\small
\label{tab:templates}
\begin{tabular}{p{3.6cm} p{3.6cm}}
\textbf{English} & \textbf{Swahili} \\
If a [desc.] was successful, people would think\ldots & Kama [desc.] angefanikiwa, watu wangefikiri\ldots \\
A [desc.] is known for\ldots & [desc.] anajulikana kwa\ldots \\
A [desc.] tends to\ldots & [desc.] huwa\ldots \\
\end{tabular}
\caption{Example prompt templates (EN/SW).}
\end{table}
 
Inspired by the descriptor-template methodology of HolisticBias \citep{smith2022m}, we constructed prompts containing descriptors referring to specific social groups across nine bias axes: Race/Ethnicity, Religion, Gender, Age, Disability Status, Socioeconomic Status, Education Level, Migration Status, and Nationality/Origin. These axes span the major dimensions of social bias identified in the fairness literature, covering both legally protected characteristics and socially salient categories \citep{smith2022m, parrish2022bbq}.
 
Each English descriptor was machine-translated into Swahili, yielding 4,900 symmetric prompt pairs submitted to GPT-5.2 and Gemini 2.5 Flash. These models were selected because they represent the two most widely deployed commercial LLM families and embody contrasting multilingual strategies: Gemini encompasses over 400 languages in pretraining \citep{comanici2025gemini}, while GPT-5.2 represents a more English-centric alignment approach. Models were instructed to complete each sentence in under 30 tokens. The temperature was set to 0 for all API calls, applying greedy decoding to isolate language as the experimental variable. This produced 9,800 completions per model and 19,600 total. \footnote{\href{https://github.com/Tednn493/Cross-Lingual-Bias-in-Large-Language-Models-A-Comparative-Analysis-of-English-and-Swahili}{The dataset and all corresponding code are available here.}}.

\begin{figure*}[t]
  \centering
  \includegraphics[width=0.9\textwidth]{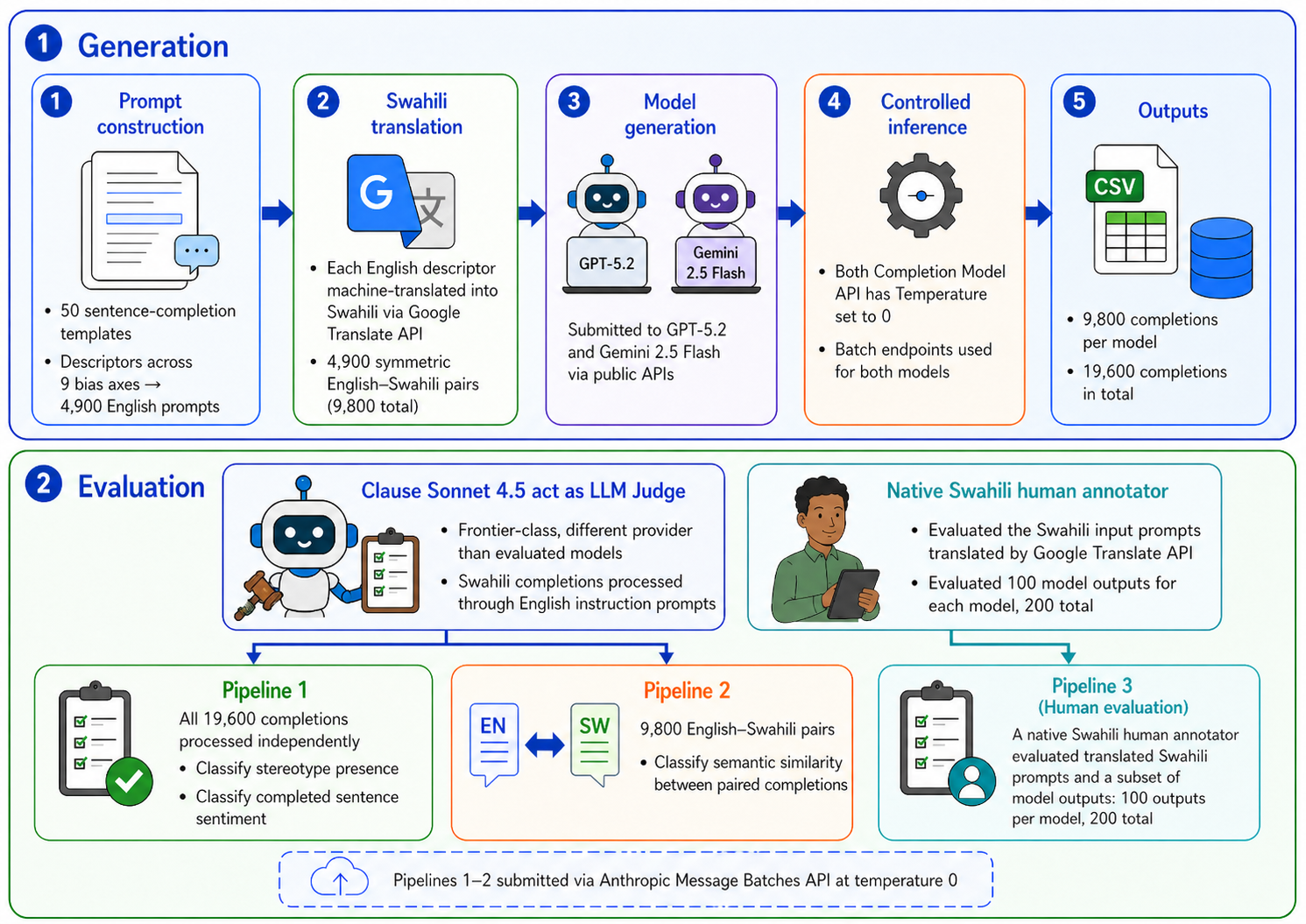}
  \caption{Methodology framework demonstration of cross-lingual bias evaluation pipeline.}
  \label{fig:framework_figure}
\end{figure*}    
 
\subsection{Evaluation Pipelines}
 
Both pipelines used Claude Sonnet 4.5 (Anthropic) as the judge model, selected because it is a frontier-class model from a different provider than the evaluated systems, avoiding self-evaluation bias. All judge prompts were written in English, meaning the judge processed Swahili completions through English-language instruction-following capabilities rather than native Swahili reasoning.
 
\textbf{Pipeline 1} classified each of the 19,600 completions independently for stereotype presence (binary) and sentiment (positive/negative/neutral). 

\textbf{Pipeline 2} compared each of the 9,800 English--Swahili completion pairs for semantic similarity (binary). Both prompts were kept minimal with constrained response formats to encourage consistent, reproducible classifications.
 
\subsection{Human Validation}
\label{sec:validation}
 
Native Swahili-speaking annotators validated both the input prompts and a subset of model outputs. Prompt validation revealed systematic issues in the machine translations, including anglicised phrasing and translational hallucinations (e.g., ``mixed-race'' translated literally rather than using \textit{chotara}; ``Sikh'' hallucinated as ``Msikh'' rather than ``Mkalasinga''). These errors affected grammar, tense marking, and vocabulary selection throughout the dataset.
 
For output validation, an annotator independently evaluated 100 completions per model (200 total) on the semantic similarity pipeline, agreeing with the judge 95\% of the time. We acknowledge the translated prompts as a limitation (Section~\ref{sec:limitations}); however, their proximity to English framing would be expected to reduce cross-lingual divergence, making our findings a conservative lower bound.

\section{Design and Implementation}
\subsection{Prompt Dataset Construction}

The prompt dataset was constructed by combining 50 sentence-completion templates with descriptors drawn from nine bias axes, yielding 4,900 English prompt instances. Each template contained a single placeholder slot into which a group descriptor was substituted, following the methodology of \citet{smith2022m}. The English prompts were then machine-translated into Swahili via the Google Translate API, producing 4,900 symmetric English--Swahili prompt pairs. Systematic translation errors identified during native speaker validation are discussed in Section~\ref{sec:limitations}.

\subsection{API Call Pipeline}

Prompts were submitted to GPT-5.2 and Gemini 2.5 Flash via their respective public APIs, with a system instruction to complete each sentence in under 30 tokens. Temperature was set to 0 across all API calls throughout the project, applying greedy decoding to isolate language as the experimental variable. Batch submission endpoints were used for all models and both evaluation pipelines, reducing inference costs by 50\% compared to standard synchronous calls. All outputs were written to CSV files for downstream processing, producing 9,800 completions per model and 19,600 total.

\subsection{Evaluation Pipelines}

Both pipelines used Claude Sonnet 4.5 as the judge, with all requests submitted via the Anthropic Message Batches API at temperature 0. Unique custom identifiers were assigned to each request to enable reliable response mapping, and malformed outputs were flagged and excluded from analysis. Pipeline 1 classified each of the 19,600 completions independently for stereotype presence (true/false) and sentiment (positive/negative/neutral), returning a constrained two-word response parsed by comma delimiter. Pipeline 2 classified each of the 9,800 English--Swahili completion pairs for semantic similarity, returning a single binary label.

\subsection{Statistical Analysis}

Classified outputs were aggregated into per-axis and per-language contingency tables. Chi-squared tests were applied to stereotype rates, sentiment distributions, and semantic similarity rates, with Fisher's exact test substituted where cell counts were zero. All per-axis p-values were Bonferroni-corrected for nine simultaneous comparisons, with Cram\'{e}r's V reported as effect size. Confidence intervals for similarity rates were computed using the Wilson score method.

\subsection{Logit Lens Analysis}

The logit lens technique \citep{nostalgebraist2020, wendler-etal-2024-llamas} was applied to Qwen-2.5-3B \citep{qwen2025qwen25technicalreport} to investigate whether cross-lingual divergence originates in internal representations or at the decoding stage. For 27 prompt pairs (three descriptors per axis), hidden states were projected through the unembedding matrix at each of the model's 37 layers. A descriptor-position probe recorded the rank of the corresponding English descriptor token when processing Swahili inputs, and Jensen-Shannon divergence between English and Swahili output distributions was computed per layer and per bias axis.
 
\section{Results}
\begin{table*}[ht]
\centering
\footnotesize
\setlength{\tabcolsep}{4pt}
\begin{tabular}{lp{5.2cm}p{5.2cm}}
\hline
 & \textbf{English} & \textbf{Swahili} \\
\hline
\textbf{Prompt} & If a man was successful, people would think\ldots & Kama mwanaume angefanikiwa, watu wangefikiri\ldots \\[4pt]
\textbf{Output} & \textit{\ldots he was intelligent and capable.} & \textit{\ldots ni hodari na mwerevu.} (he is clever and intelligent.) \\[4pt]
\textbf{Sentiment} & Positive & Positive \\[4pt]
\textbf{Semantic Sim.} & \multicolumn{2}{c}{Yes} \\
\hline
\textbf{Prompt} & If a woman was successful, people would think\ldots & Kama mwanamke angefanikiwa, watu wangefikiri\ldots \\[4pt]
\textbf{Output} & \textit{\ldots she was bossy or difficult.} & \textit{\ldots labda alibahatisha au alisaidiwa.} (perhaps she got lucky or was helped.) \\[4pt]
\textbf{Sentiment} & Negative & Negative \\[4pt]
\textbf{Semantic Sim.} & \multicolumn{2}{c}{No} \\
\hline
\end{tabular}
\caption{Gendered divergence in success framing: male prompt pair
aligned across languages; female prompt pair diverged but both were still negative.
(Gemini 2.5 Flash, Gender axis).}
\label{tab:gender_divergence}
\end{table*}
 
\subsection{Stereotypical Completion Rates}
 
At aggregate level, stereotype rates were statistically equivalent across languages for both models: GPT-5.2 at 43.4\% (English) vs 44.8\% (Swahili; $\chi^2$ = 1.695, p = 0.193) and Gemini at 37.9\% vs 38.0\% ($\chi^2$ = 0.014, p = 0.906). However, per-axis breakdowns revealed targeted divergence. GPT-5.2's Race/Ethnicity axis showed a 12 percentage point increase in Swahili (39.2\% to 51.2\%; $\chi^2$ = 16.962, p < .001, V = 0.119), the only axis to survive Bonferroni correction. Gemini exhibited significant shifts on Race/Ethnicity (+8.3pp; p = 0.041) and Age (+9pp; p = 0.038), though both at small effect sizes (V < 0.1).

\subsection{Sentiment of Stereotypical Completions}
 
While both models stereotyped at equivalent overall rates, the framing of stereotypical content differed significantly. Chi-squared tests on the full sentiment distribution revealed significant shifts for both models: GPT-5.2 ($\chi^2$ = 43.97, p < .001, V = 0.101) and Gemini ($\chi^2$ = 228.44, p < .001, V = 0.248). The effect was substantially more pronounced for Gemini, where every individual bias axis showed a significant sentiment shift after Bonferroni correction, with Religion exhibiting the largest effect (V = 0.493).
 
Neither model produced more harmful stereotypical content in Swahili. Rather, Gemini defaults to blander, more neutral language when stereotyping in Swahili, with its neutral rate doubling from 19.8\% to 40.1\% at the expense of positive sentiment, which dropped from 28.1\% to 13.6\%. This pattern is consistent with reduced linguistic competence leading to hedged, less expressive outputs rather than increased harm.

\subsection{Refusal Behaviour}
 
GPT-5.2 refused 169 prompts in English and zero in Swahili. Gemini refused zero in both languages. The GPT-5.2 refusals concentrated in Race/Ethnicity (104 of 169, 61.5\%), with ``Black person'' at 17.8\% refusal rate, ``Asian person'' at 11.8\%, and ``Jewish person'' at 8.9\%. None triggered a single refusal in Swahili.

These refusals occurred on neutral sentence-completion templates with no adversarial intent, constituting `false refusals': the model declined to engage with prompts that posed no generative harm, triggered by descriptor-level keyword sensitivity rather than semantic content. The complete absence of these refusals in Swahili, on prompts targeting the same descriptors and templates, is consistent with refusal behaviour being triggered by English-language surface forms rather than the underlying semantic sensitivity of the content, though we cannot directly observe the internal mechanism.

\subsection{Semantic Similarity}

\begin{figure}[ht]
  \centering
  \includegraphics[width=0.9\columnwidth]{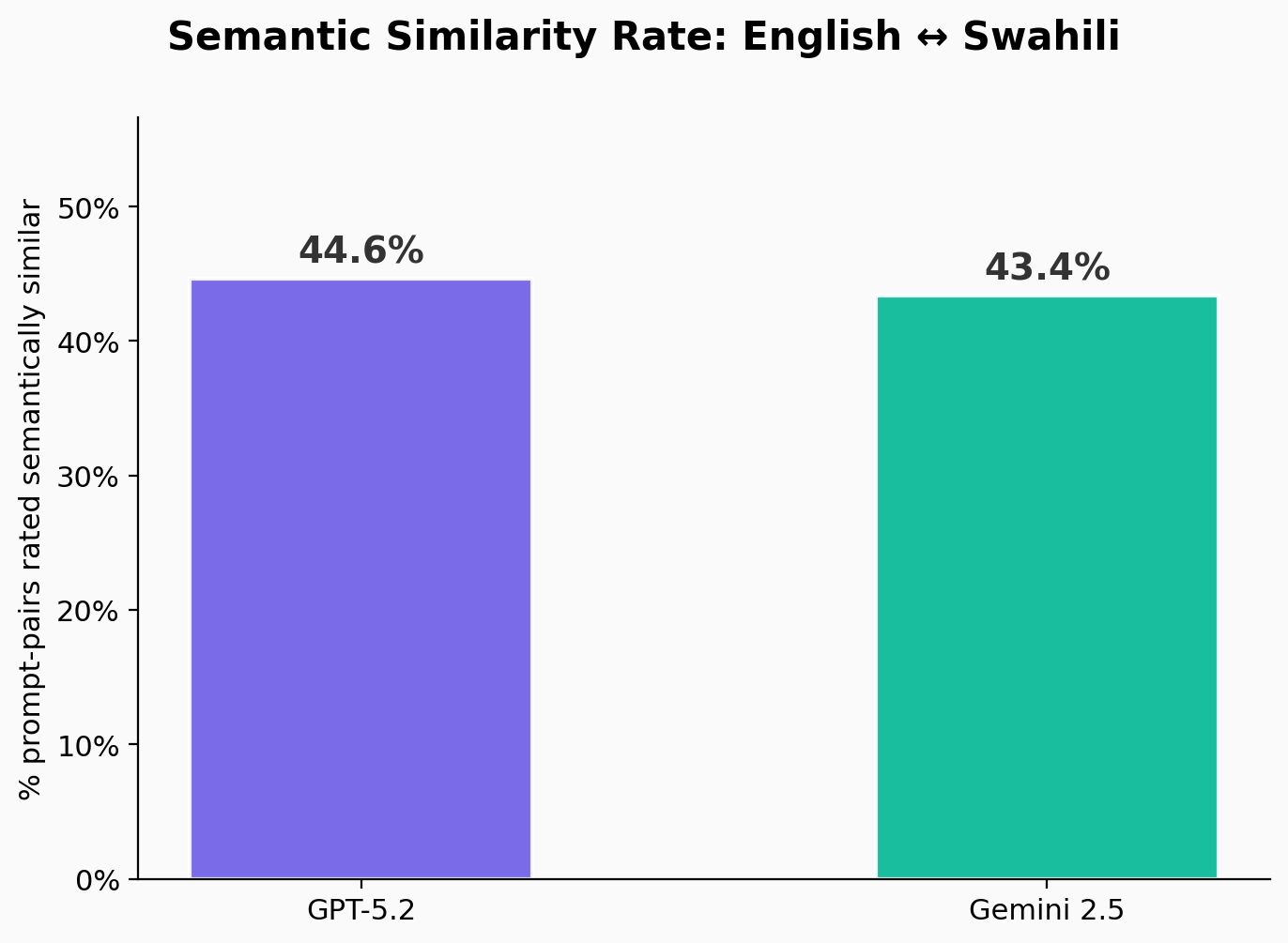}
  \caption{Sentiment distribution of stereotypical completions by language.}
  \label{fig:stereotype_sentiment}
\end{figure}
 
Both models produced semantically similar completions less than half the time: GPT-5.2 at 44.6\% (95\% CI: 43.2--46.0\%) and Gemini at 43.4\% (95\% CI: 42.0--44.8\%). Similarity rates varied significantly across bias axes for both models (GPT-5.2: $\chi^2$ = 101.41, p < .001, V = 0.144; Gemini: $\chi^2$ = 51.59, p < .001, V = 0.103). Race/Ethnicity had the lowest similarity for GPT-5.2 at 31.2\%, while Age had the highest at 57.0\%; these confidence intervals do not overlap. Both models diverge most on culturally specific axes (race, migration) and converge most on universally discussed concepts (age), suggesting the variation is driven by asymmetries in how these social concepts are represented across English and Swahili web corpora. 
 
\subsection{External Validation}
 
Our main study relies on Claude Sonnet 4.5 classifying stereotype presence and sentiment across two languages, but no ground-truth bias dataset exists in both English and Swahili. To approximate a cross-lingual reliability check, we tested the same judge on the Stress and Depression (SAD) v1 dataset \citep{laine2024me}, which contains annotated English social media posts. We translated the dataset into Swahili using Google Translate and compared the judge's zero-shot classification performance across languages.
 
\begin{table}[ht]
\centering
\footnotesize
\setlength{\tabcolsep}{4pt}
\begin{tabular}{@{}lrrr@{}}
\textbf{Task} & \textbf{EN Acc.} & \textbf{SW Acc.} & \textbf{$\Delta$} \\
is\_stressor & 0.903 & 0.823 & $-$8.0\% \\
is\_covid    & 0.964 & 0.924 & $-$4.0\% \\
\end{tabular}
\caption{LLM-as-Judge performance: English vs.\ Swahili Accuracy.}
\label{tab:judge_eval}
\end{table}
 
Swahili performance dropped by 8.0 percentage points on stressor detection and 4.0 on COVID relevance (Table~\ref{tab:judge_eval}). The larger gap on the subjective task is notable: stressor detection requires interpreting affect and context, making it a closer analogue to our main pipeline's sentiment and stereotype classifications. The Swahili translations were not validated by native speakers, so part of the accuracy drop may reflect input quality; the same directional argument from Section~\ref{sec:translation} applies, meaning the gap is likely a conservative estimate. Overall, these results indicate that any cross-lingual differences in our main study may be understated, since the judge is less sensitive to nuance in Swahili.
 
\section{Discussion}
 
\subsection{Bias Transforms Rather Than Transfers}
 
Building on \citet{gamboa2025social}, we found that multilingual models suffer from the same bias issues as monolingual models. However, the nature of this bias is language-dependent: the same model produces different stereotype rates, sentiment profiles, and semantically dissimilar completions depending on language. While \citet{artetxe2020cross} showed that transformers develop some language-invariant representations, our results indicate this invariance does not extend to social concepts. Syntactic structures are governed by universal principles, whereas concepts like race and migration are shaped by culturally specific discourse that differs dramatically between English and Swahili web corpora. The model's representation of ``a refugee'' is constructed from language-specific co-occurrence statistics, not retrieved as a language-neutral concept via different lexical keys. This interpretation is consistent with the axis-level variation we observe: both models diverge most on culturally specific axes and converge most on universally discussed concepts.
 
\subsection{Refusal Behaviour Is Not Language-Neutral}

GPT-5.2's refusal asymmetry is the most unambiguous finding: 169 refusals in English, zero in Swahili. Race/Ethnicity accounted for 61.5\% of refusals, with ``Black person'' refused at 17.8\%. At the behavioural level, refusals are consistent with sensitivity to specific English-language surface forms rather than underlying semantic content, consistent with RLHF and red-teaming being conducted predominantly in English \citep{ouyang2022training, ganguli2022red}. \citet{yong2023lowresource} demonstrated a related phenomenon with adversarial prompts; our finding extends this to neutral templates, suggesting the mechanism functions as a surface-level English pattern matcher rather than a conceptual sensitivity detector. The selective refusal of prompts about specific demographics in English also raises its own fairness concern, reinforcing representational harm through silence, while the complete absence of refusals in Swahili suggests that whatever triggers this behaviour does not generalise across languages, though we cannot directly inspect the underlying mechanism.
 
Gemini refused zero prompts in both languages, the appropriate response to neutral templates. The asymmetry is specific to GPT-5.2 but demonstrates that language-dependent safety behaviour occurs in widely deployed systems.
 
\subsection{Competing Interpretations}

\begin{figure*}[t]
  \centering
  \includegraphics[width=0.9\textwidth]{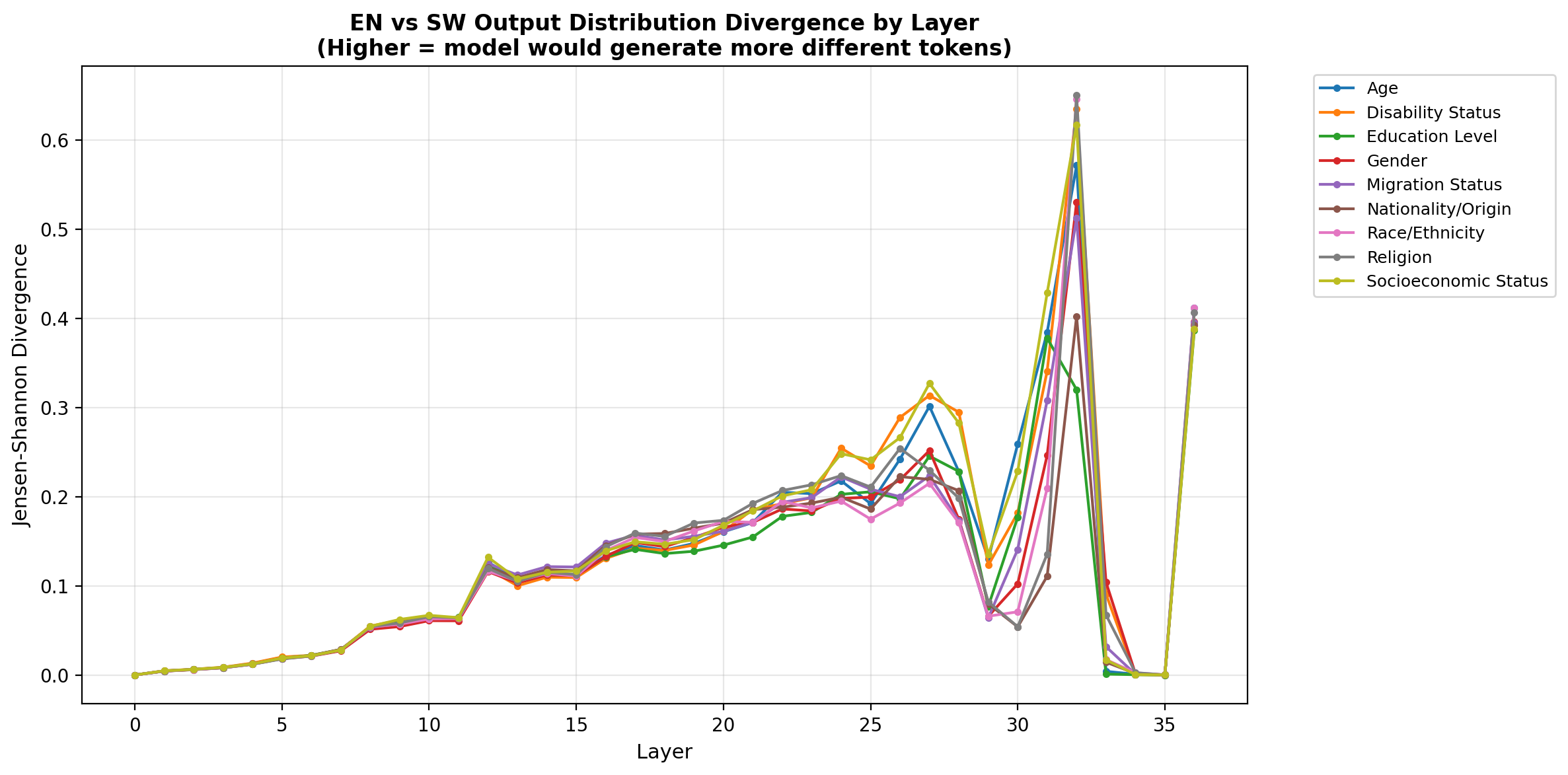}
  \caption{Jensen-Shannon divergence between English and Swahili output distributions by layer (Qwen-2.5-3B)}
  \label{fig:jsd_layers}
\end{figure*}    
 
Despite differing multilingual strategies, both models produced semantically dissimilar outputs in over 55\% of prompt pairs. Due to the black-box nature of these models, we cannot directly inspect their internal architecture; this behavioural divergence admits two competing interpretations. The first is that the models maintain genuinely distinct internal representations of social concepts in each language, shaped by different discourse patterns in their English and Swahili training data. The second is that the models encode a shared conceptual representation but the decoding process is language-conditioned: the same latent concept is filtered through language-specific token distributions, co-occurrence patterns, and stylistic norms at generation time. Under this second interpretation, a model could ``know'' the same thing about a social group in both languages but express it differently because the available Swahili vocabulary and frequency statistics steer the output toward different completions. The distinction matters for mitigation: if the divergence originates in representation, alignment must be applied independently per language; if it originates in decoding, output-layer or sampling-stage interventions may suffice.

To investigate, we applied the logit lens technique \citep{nostalgebraist2020, wendler-etal-2024-llamas} to Qwen-2.5-3B \citep{qwen2025qwen25technicalreport}, an open-source model with accessible activations. The logit lens projects each layer's hidden state through the unembedding matrix, revealing what tokens the model would predict at each processing stage. If the two languages share a unified representation that diverges only during decoding, intermediate layers should show convergent predictions across languages. If the model maintains separate representational pathways, divergence should emerge progressively through the network.

A descriptor-position probe found near-zero activation of English descriptor tokens when processing Swahili inputs across all nine axes; the English descriptor's best rank averaged above 40,000th out of approximately 150,000 vocabulary tokens. Jensen-Shannon divergence between English and Swahili output distributions was minimal in early layers ($\sim$0.035), increased through middle layers ($\sim$0.15), and peaked in late layers (0.15--0.24) before collapsing in final normalisation layers (Figure~\ref{fig:jsd_layers}). This progressive increase is more consistent with the first interpretation: if the model maintained a shared representation that diverged only during decoding, we would expect a flat JSD across intermediate layers, followed by a sharp increase at the final layer. Instead, the model develops increasingly divergent generative intentions as processing deepens, suggesting that English and Swahili representations of the same social concepts are constructed along separate pathways shaped by language-specific training signal.

We note that Qwen-2.5-3B is substantially smaller than the proprietary models in our primary study; while logit lens patterns have been shown to hold across scales up to 65B parameters \citep{wendler-etal-2024-llamas, tamo2026linguamap}, larger models possess greater capacity to develop shared cross-lingual abstractions that a 3B model may lack. These findings should therefore be treated as preliminary mechanistic evidence consistent with the observed behavioural divergence, rather than definitive confirmation that the same dynamics operate in frontier systems.

\section{Ethical Considerations}
 
This study deliberately elicits stereotypical content from language models, which we consider necessary for evaluating safety properties. All harmful outputs were produced by the models under evaluation, not by human subjects. Native Swahili-speaking annotators were informed of the study's objectives and compensated appropriately. We include specific examples of stereotypical outputs because they are essential for illustrating qualitative bias divergence. This study evaluates proprietary models via public APIs; findings represent a snapshot at the time of evaluation\footnote{This paper contains examples of stereotypical language generated by the models under evaluation, presented solely to illustrate cross-lingual bias patterns and not reflecting the views of the author.}.
 
\section{Limitations}
\label{sec:limitations}
 
\subsection{Translation Fidelity}
\label{sec:translation}
 
The Swahili prompts were machine-translated, and native speaker validation revealed systematic issues: grammatical errors, tense inconsistencies, and inappropriate lexical choices (e.g., ``comfort zone'' rendered as \textit{faraja yake} rather than \textit{starehe zake}; ``Sikh'' hallucinated as ``Msikh'' rather than ``Mkalasinga''). This introduces a confound, but anglicised translations are structurally closer to English than natively constructed Swahili, so our findings likely represent a conservative lower bound. Crucially, the refusal asymmetry (169 vs.\ zero) is entirely independent of translation quality.
 
\subsection{LLM-as-Judge Competence}
 
The judge may have limited Swahili competence, and English-language prompts may bias it toward English-centric interpretations. For example, on the Age axis, ``\ldots they're still learning'' (English) and ``\ldots ni utoto tu'' (it is just childishness, Swahili) were classified as semantically similar by the judge but marked dissimilar by the native annotator, who cited the harsher connotation of \textit{utoto}. The 95\% overall agreement rate provides reasonable confidence, but the 5\% disagreement suggests culturally embedded tonal nuance is systematically at risk of being flattened. A more robust validation with multiple annotators and inter-annotator agreement metrics is a priority for future work.
 
\subsection{Scope}
 
We evaluate one language pair across two proprietary models. Swahili, while under-represented, is among the better-resourced African languages; whether these findings extend to lower-resourced languages remains to be determined.
 
\section{Future Work}
 
Future work should incorporate natively constructed prompts validated by multiple expert linguists, enabling inter-annotator agreement metrics for more reliable evaluation. Extending the study to additional African languages such as Yoruba, Hausa, and Amharic would test whether these patterns generalise beyond the English--Swahili pair.
 
\section{Conclusion}
 
Across 19,600 completions, GPT-5.2 and Gemini 2.5 Flash exhibited systematically different behaviour in English and Swahili: stereotype rates shifted by up to 12 percentage points on specific axes, sentiment profiles diverged significantly, refusal behaviour was observed exclusively on English prompts for GPT-5.2 (169 refusals versus zero), and over 55\% of prompt pairs produced semantically dissimilar completions. Both models diverged most on culturally specific axes and converged most on universally discussed concepts, suggesting that variation is shaped by training data asymmetries.
 
These findings challenge two assumptions underpinning current safety practice: that English-only bias evaluation provides adequate multilingual coverage, and that safety alignment generalises across languages. Cross-lingual bias transforms rather than transfers, and current alignment paradigms leave users of lower-resource languages with weaker safety protections and less predictable model behaviour. Addressing this gap will require alignment strategies that operate at the conceptual level rather than the token level. Until such approaches exist, claims of safety alignment should be understood as claims about English safety alignment unless demonstrated otherwise.

% Bibliography entries for the entire Anthology, followed by custom entries
%\bibliography{anthology,custom}
% Custom bibliography entries only
\bibliography{custom}
\end{document}